\documentclass[sigconf]{acmart}

\usepackage{booktabs} 
\newcommand{\bbR}{\mathbb{R}}

\newcommand{\calX}{\mathcal{X}}
\newcommand{\calL}{\mathcal{L}}

\newcommand{\calZ}{\mathcal{Z}}
\newcommand{\calN}{\mathcal{N}}

\newcommand{\calS}{\mathcal{S}}

\newcommand{\bx}{\mbox{$\mathbf{x}$}}
\newcommand{\bz}{\mbox{$\mathbf{z}$}}
\newcommand{\bC}{\mbox{$\mathbf{C}$}}
\newcommand{\bJ}{\mbox{$\mathbf{J}$}}

\newcommand{\bM}{\mbox{$\mathbf{M}$}}
\newcommand{\bX}{\mbox{$\mathbf{X}$}}
\newcommand{\bU}{\mbox{$\mathbf{U}$}}
\setcopyright{acmcopyright}

\acmDOI{10.475/123_4}

\acmISBN{123-4567-24-567/08/06}

\acmConference[ICVGIP'18]{ICVGIP}{December 2018}{Hyderabad, India}
\acmYear{2018}
\copyrightyear{2016}

\acmPrice{15.00}

\acmSubmissionID{296}

\begin{document}
\title{Geometry of Deep Generative Models for Disentangled Representations}

\author{Ankita Shukla}
\orcid{}
\affiliation{%
  \institution{IIIT-Delhi}
  \streetaddress{P.O. Box 1212}
  \city{Delhi}
  \state{India}
  \postcode{43017-6221}
}
\email{ankitas@iiitd.ac.in}

\author{Shagun Uppal}
\affiliation{%
  \institution{IIIT-Delhi}
  \streetaddress{P.O. Box 1212}
 \city{Delhi}
  \state{India}
  \postcode{43017-6221}
}
\email{shagun16088@iiitd.ac.in}

\author{Sarthak Bhagat}
\affiliation{%
  \institution{IIIT-Delhi}
  \streetaddress{P.O. Box 1212}
  \city{Delhi}
  \state{India}
  \postcode{43017-6221}
}
\email{sarthak16189@iiitd.ac.in}

\author{Saket Anand}
\affiliation{%
  \institution{IIIT-Delhi}
  \city{Delhi}
  \state{India}
  \postcode{43017-6221}
}
\email{anands@iiitd.ac.in}

\author{Pavan Turaga}
\affiliation{%
  \institution{Arizona State University}
  \streetaddress{P.O. Box 1212}
  \city{Tempe}
  \state{Arizona, USA}
  \postcode{43017-6221}
}
\email{pturaga@asu.edu}

\renewcommand{\shortauthors}{A. Shukla et al.}

\begin{abstract}
Deep generative models like variational autoencoders approximate the intrinsic geometry of high dimensional data manifolds by learning low-dimensional latent-space variables and an embedding function. The geometric properties of these latent spaces has been studied under the lens of Riemannian geometry; via analysis of the non-linearity of the generator function. In new developments, deep generative models have been used for learning semantically meaningful `disentangled' representations; that capture task relevant attributes while being invariant to other attributes. In this work, we explore the geometry of popular generative models for disentangled representation learning. We use several metrics to compare the properties of latent spaces of disentangled representation models in terms of class separability and curvature of the latent-space. The results we obtain establish that the class distinguishable features in the disentangled latent space exhibits higher curvature as opposed to a variational autoencoder. We evaluate and compare the geometry of three such models with variational autoencoder on two different datasets. Further, our results show that distances and interpolation in the latent space are significantly improved with Riemannian metrics derived from the curvature of the space. We expect these results will have implications on understanding how deep-networks can be made more robust, generalizable, as well as interpretable.
\end{abstract}  

%
%


\maketitle
\section{Introduction}
Deep learning approaches have achieved state-of-the-art results in many applications, drawing on the strength of large-scale hierarchical architectures to capture the complex structure of image data. Despite the great success of deep networks in various tasks \cite{he_cvpr2016,Gregor_ICML2015}, the learned representations are generally opaque, and limited in their generalization and interpretability. To address these limitations, various approaches have been developed that try to learn meaningful representations that can generalize well, are robust across data variations, and have some interpretability. The family of such representations is known as disentangled representations \cite{Bengio_2013}, whose goal is to separate (disentangle) the different salient factors or attributes that give rise to variation in data samples. 

On the other hand, a lot of work in computer vision and graphics has focused on developing analytical forward (and inverse) models of image formation, written as combinations of various factors such as pose, shape, illumination, reflectance, inter-reflections, shading, etc. These variables often interact in highly non-linear ways, partly due to the intrinsic non-Euclidean nature of the space in which these variables reside. Common assumptions to simplify the generation model include assuming Lambertian reflectance properties, near-convex object shapes etc. Several results have been developed which under different assumptions and different combinations of factors show that the data manifold generated is in fact non-linear (c.f. \cite{Georghiades2001,Xu2007}). The non-linearity in the observed data manifold is due to both the intrinsic non-linearity of certain factors of variation (like pose), and the non-linearity in the forward model itself (due to illumination and shading for instance). 

In newer developments, various disentangled representation learning approaches \cite{Reed_ICML2014,Cheung2014,Hinton2011,Goroshin2015,Yang_NIPS2015} have been proposed. The aim of these approaches is to learn feature representations that reflect the  underlying semantics for the specified factor while being invariant to other common factors. For example, in the task of human face recognition, the features for recognizing the face based on identity in an image should be invariant to factors like pose, illumination, or expression in the image. The application of such representations not only benefits standard tasks like classification and recognition, but also tasks like image generation and synthesis with specific attributes.


The most widely followed paradigm in learning such representations is to achieve a decomposition of the feature space where different dimensions correspond to disjoint factors of variation. Such a representation allows one to generate novel images where only a specified factor is changed while other factors of variations are kept fixed. In the context of this paper, we are specifically looking at disentangled representations that factorize the latent space into two components \cite{Szabo_2017,Jha_2018,Mathieu_2016}, where one captures the task relevant attribute while all other factors are considered in the other component. We refer to these components as specified and unspecified components of the representation respectively. As an illustrative example for face recognition, the model shown in the figure \ref{fig: model}, has identity as the specified factor and all other variables like pose, illumination, facial hair as the unspecified component. 

As most recent representation learning approaches use neural networks as the base models, there has been an increased interest in understanding their geometry. The work of Brahma et al. \cite{Brahma2016} focuses on providing theoretical insights and justification behind the success of deep learning models. Following the success of deep generative models like Generative Adversarial Networks (GANs) and Variational Autoencoders (VAEs) in various applications like realistic image generation \cite{DCGAN_2015}, learning interpretable features \cite{Szabo_2017,Mathieu_2016} and unsupervised domain adaptation \cite{DANN}, several recent works have focused on establishing the geometry of generative models \cite{Hauberg_2017,Kuhnel2018,Shao2018,Laine2018}. Shao et al. \cite{Shao2018} suggest that the generated Riemannian data manifold generated by a VAE, while nonlinear has very low curvature. This conclusion further implies that the linear movements in the latent space result in movements approximately along the geodesics of the generated data manifold. 
Thus, statistics computed on a much smaller dimensional latent space, can be used along with the Riemannian geometry induced by the generator mapping function $f(Z)$, to draw inferences in the high dimensional data manifold. Later, \cite{Hauberg_2017,Kuhnel2018} showed that the latent space is typically nonlinear and computations should appropriately account for its curvature. The analysis in these papers has been restricted to latent space of one type of generative model \emph{i.e.} VAEs. 

The generator of a deep generative model like VAE can be seen as a mapping from the low dimensional latent space to the data manifold embedded in a much higher dimensional space, which in turn permits us to define the Riemannian metric of the data manifold via the Jacobian of the mapping. The work of Shao et. al. \cite{Shao2018} develops the Riemannian metric for deep generative models and analyzes the curvature of the generated data manifold. They empirically show that for VAEs, the resulting manifold is non-linear, yet has a low curvature as indicated by the low disparity between the Euclidean and the geodesic distances. However, they do not directly provide a compact measure for quantifying the curvature. Aravanitdis et. al. \cite{Hauberg_2017} also use the Riemannian metric to define non-linear statistics in the latent space of VAEs, and argue that the latent space is a curved space and advance the use of geodesic interpolation.   



As pointed out by these recent works, leveraging the Riemannian metric of generative models helps in smoother interpolations as well as more meaningful interpretation of distances as opposed to the default Euclidean metric. 
As in the previous works, we restrict our study to deep generative models, however, we focus our investigation on the geometric properties of VAE-based disentangled latent space models. Furthermore, we aim to quantify the curvature of the latent space and resort to different metrics for studying the following. Firstly, we validate and quantify the near zero curvature of VAE as addressed in \cite{Shao2018}. Secondly, we extend the geometric analysis to disentangled representation learning models and quantify the effect of class separability on the curvature of the generated data manifold.




The organization of the paper is as follows. In Section \ref{sec: back} we discuss the relevant background of generative models and disentangled representation learning required for understanding of this work, 
followed by the different metrics used to establish geometrical insights in Section \ref{sec: geometry}. Lastly, we present the experiments and results in Section \ref{sec: exp} followed by conclusion in Section \ref{sec: con}.
\begin{figure*}
	\begin{center}
		\includegraphics[trim={0 1.5cm 0 3cm}, clip= true, width =0.9\textwidth]{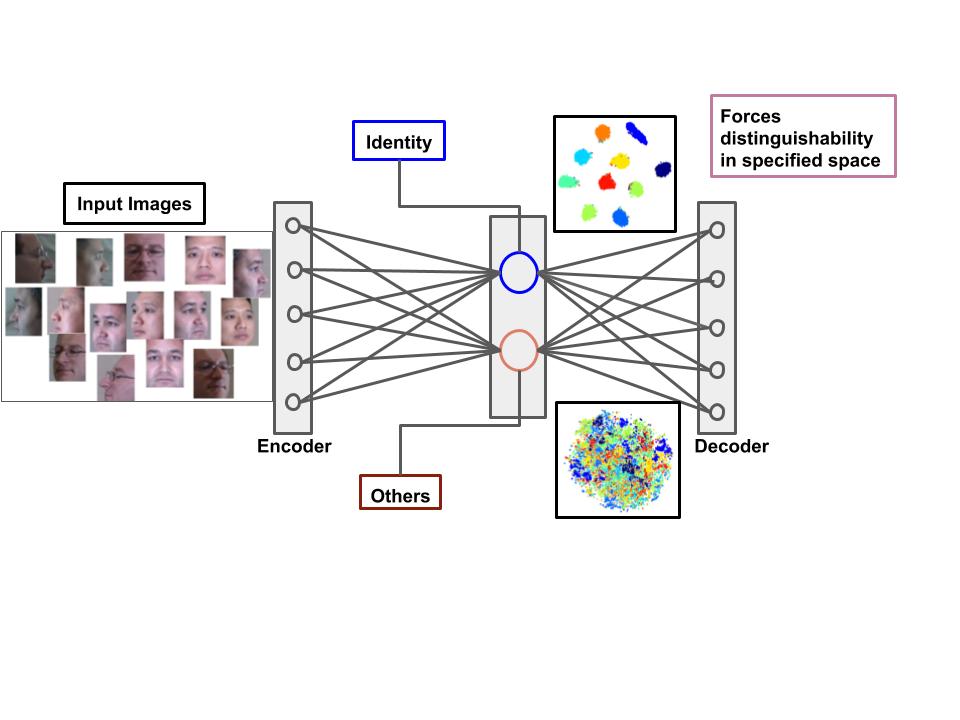}
        \end{center}
        \vspace{-3cm}
\caption{Model for learning factorized latent space representation for human face recognition. The identity constitutes the specified component while all other factors such as pose and illumination are considered in unspecified space }
\end{figure*}\label{fig: model}



\section{Background}\label{sec:  back}
In this section, we discuss the necessary background of generative models and disentangled representation learning model relevant in the context of our work.
\subsection{Notation}
We denote the samples from the data manifold $\calX $ as $\bx_1,\bx_2, \ldots \bx_n \in \bbR^n$. The encoder, decoder and discriminator are denoted by $Enc$, $Dec$ and $Dsc$ respectively. The latent space variables for specified and unspecified components are given by $s$ and $z$ respectively.  The encoder mapping functions are $g_s: \calX \rightarrow \calS$ and $g_z: \calX \rightarrow \calZ$ for specified and unspecified components respectively. Similarly, the decoder or generator functions are given as $(f_s,f_z): (\calS,\calZ) \rightarrow \calX$. The encoder and decoder parameters are denoted by $\theta$ and $\phi$ respectively. 
\subsection{Generative Models}
Deep generative models learn to approximate the data distribution by modeling the latent variables and a generator function that maps the latent space variables to the data manifold. For a given set of data samples $\bx \in \calX$, we denote the corresponding latent space variables as $\bz \in \calZ$, the generator mapping as $f: \calZ \rightarrow \calX$, such that $\bx = f(\bz)$. This mapping from the latent space is an embedding of coordinates $\calZ$ to the data manifold $\calX$, with certain network requirements \cite{Shao2018}. The two generative models used in this paper are: Variational Autoencoders (VAEs) and Generative Adversarial Networks (GANs)

\textbf{VAEs} \cite{kingma}
Variational autoencoder is a generative model comprising of an encoder network that maps the data manifold to low dimensional latent space, and a decoder network that learns to map these representations back to the data manifold.  The latent space $\calZ$ is constrained such that the $\bz$ are sampled from a specific probability distribution. Most popularly, the prior distribution is constrained to be the Gaussian $p(\bz) = \calN(0,{I}_d)$. 
Therefore, in case of VAE, the generator mapping $f$ parameterizes the data distribution and the posterior distribution $q_{\theta}(z|x)$ parametrizes the encoder, constrained to be a Gaussian and the unknown posterior distribution term $p_{\phi}(x|z)$ defines the decoder. Here, $\theta$ and $\phi$ are encoder and decoder parameters respectively. The optimal $\theta$ and $\phi$ are obtained by maximizing the lower bound of marginal likelihood $p(\bx)$ as 

\begin{align}
\arg \max_{\theta,\phi} \mathbb{E}_{q_{\phi}(\bz|\bx)}[log(p_{\theta}(\bx|\bz))]-\textbf{KL}(q_{\phi}(\bz|\bx)||p(\bz))
\end{align}

\textbf{GANs} \cite{GANs_Goodfellow_NIPS2014} are generative models with two competing neural networks: a discriminator and a generator that together learn the data distribution in a minmax game. The discriminator tries to distinguish between the samples generated from the data distribution from the noise samples while the generator gradually learns to generate samples that can fool the discriminator. 

\subsection{Disentangled Representation Learning}
The focus of this work is limited to factorized disentangled representations that partition the latent space into two codes: one code for the specified factor and the other code for all uncontrolled or unspecified factors. For example, the latent space of face data set can be partitioned with identity as specified factor whereas factors like illumination, pose and expression together as unspecified factor. In this work, we study the geometry of three models proposed for disentangling \cite{Szabo_2017}, \cite{Mathieu_2016} and \cite{Jha_2018}.

\textbf{Szabo \emph{et al.}\cite{Szabo_2017}}
The model learns a disentangled latent representation using an encoder-decoder architecture with weakly labeled data in the form of triplets $ \{\bx_1,\bx_2,\bx_3\} $, where $ \bx_1 $ and $ \bx_2 $ have the same label, while $ \bx_3 $ has a different label. Note that \emph{absolute} labels are not required for this architecture, hence the weaker form of supervision. As shown in figure 2, during the training, the specified and unspecified components are swapped and an appropriate loss function is optimized to ensure that the decoder learns to generate realistic samples. In the first step, since $ \bx_1 $ and $ \bx_2 $ have the same labels, the encoder should learn to satisfy $ s_1\simeq s_2 $. Upon swapping the unspecified components ($ z_1 \text{ and } z_2$), a simple $ \ell_2 $ loss suffices to ensure that the generated sample $ \bx_{2\oplus 1} $ is similar to $ \bx_2 $ as well as $ \bx_{1\oplus 2} $ looks similar to $ \bx_1 $. On the other hand, when $ z_3 $ and $ z_1 $ (or $ z_2 $) are swapped, the loss cannot be an $ \ell_2 $-norm, which Szabo et al. \cite{Szabo_2017} circumvent by using a discriminator with an adversarial loss. The discriminator is trained over real pairs ($ \bx_1,\bx_2 $) and fake ones ($\bx_{3\oplus 1}, \bx_1 $), so as to enable the decoder to generate samples that resemble those from the distribution of $ \bx_1 $ and $ \bx_2 $. 


The objective function consist of two terms: autoencoder loss and adversarial loss, and is given by 
\begin{align}
\min_{Dec, Enc} \max_{Dsc} \calL_{AE}(Dec, Enc)+\lambda \calL_{GAN}(Dec, Enc, Dsc)
\end{align}
Here $\calL_{AE}$ is given by 
\begin{align}
\calL_{AE} = \mathbb{E}_{\bx_1,\bx_2}||\bx_1-f(g_s(\bx_1),g_z(\bx_2))||_2^2 + ||\bx_2-f(g_z(\bx_2),g_s(\bx_1))||_2^2
\end{align}
 and, the adversarial loss is given by 
 \begin{align}
 \calL_{GAN} = \mathbb{E}_{\bx_1,\bx_2}[\log(d(\bx_1,\bx_2))]+  \mathbb{E}_{\bx_1,\bx_3}[\log(1-d(\bx_1,\bx_{}))]
 \end{align}
\begin{figure}
	\begin{center}
		\includegraphics[trim={0 0 0 0},clip,width =0.4\textwidth]{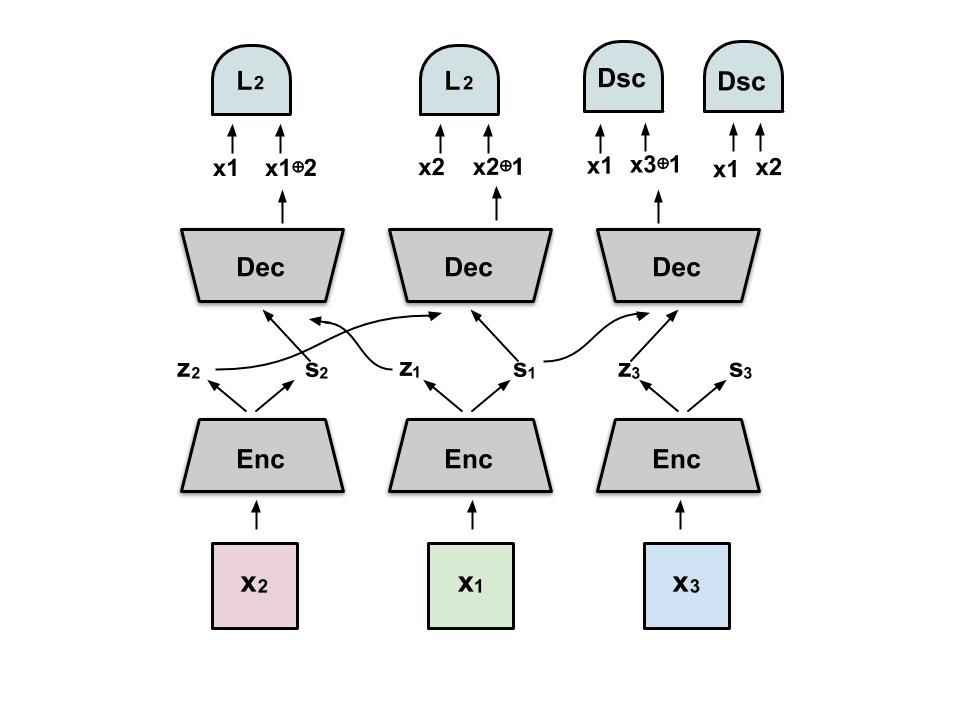}
        \end{center}
 \caption{Network for Disentangled Representation Learning given in \cite{Szabo_2017}.}
\end{figure}\label{fig: ch_model}

The model by \textbf{Mathieu \emph{et al.}} \cite{Mathieu_2016} is similar to the one described above, however, the key difference is they train the unspecified component as a VAE, i.e., impose a KL divergence term such that the unspecified component has a standard normal distribution. The work by \textbf{ Jha \emph{et al.}} \cite{Jha_2018} improves upon the two approaches \cite{Mathieu_2016,Szabo_2017} by substituting the adversarial training strategy with cycle-consistency of the unspecified latent space.   \\


\noindent \textbf{\textit{Geometric Perspective}} All the three methods, despite their differences, attempt to partition the space into specified and unspecified components. In context of the work in this paper, the two important and common aspects of these methods are
\begin{itemize}
\item Encoder-decoder based architectures, that partition the latent space into specified and unspecified factors. 
\item Specified latent space comprises discriminative features of the specified attributes, while the unspecified space contains uninformative, nuisance variables. 
\end{itemize}
We focus on analyzing the geometry of the two components of the learned latent space ($ s,z $) and the corresponding generator mapping functions ($ fs(\cdot),f_z(\cdot) $). We expect this analysis and quantitative validation to provide insights into the geometric aspects of learning disentangled latent spaces, which in-turn may help design better network architectures and training strategies. 



\section{Geometry of Latent Space of Factorized Representations}\label{sec: geometry}
In this section, we present the different metrics used to compare the latent spaces of the learned disentangled representation.
 \subsection{Euclidean vs Riemannian Metric}
The latent space of generative models provide a low dimensional representation of the data manifold via nonlinear functions implemented by the encoder and decoder. Thus statistical computations based on manifold theory is more appropriate as opposed to the Euclidean space assumption. As pointed out by works in \cite{Hauberg_2017,Kuhnel2018}, the latent space is the coordinates for the data manifold though a generator mapping function. Thus distances and other statistics are better defined with a Riemannian metric. This manifold assumption holds true provided the generator is a smooth function.

In deep generative models, the network is a composition of multiple convolutonal layers followed by activation layers that bring in the nonlinearity of the feature space. In order to obtain a smooth generator function, the activation function such as ELU (exponential linear unit) is used over the more popular ReLU (rectified linear unit).

Thus, the two conditions required for the generator mapping $f$ to define a smooth manifold \cite{Shao2018} are:
\begin{itemize}
\item The activation function is smooth and monotonic function. 
\item The weight matrices for the layers are full rank.
This condition in effect translates to the full rank condition of  Jacobian matrix at every point in the latent space
\\
\qquad $rank(\bJ_{f}(\bz)) = d$
\\
where $z$ is the point in the latent and $d$ is the dimension of the latent space.
\end{itemize}

Provided that the above conditions are satisfied, every point in the coordinate latent space is mapped uniquely on to the data manifold. This connection allows one to perform operations in the low dimensional latent space, that is relatively computationally cheap as opposed to their higher dimensional counterpart, the data manifold. Given a point in the latent space, the Jacobian provides a mapping from the tangent space of latent space to the tangent space of the data manifold. The associated Jacobian at every point in the coordinate space allows to define a local metric at every point in the space that accounts for distortion brought by low dimensional latent space representations.
Thus, the Riemannian metric is defined at every point  $\bz$ as a symmetric positive definite matrix $\bM_{\bz}$ 
as 
\begin{align}
\bM_{\bz} = \bJ_{\tiny f}(\bz)^\top\bJ_{\tiny f}(\bz)
\end{align}
Here, $\bJ_{f}(z)$ is the Jacobian matrix at point $\bz$ and $f()$ is the generator function. 
\subsection{Residual Normalized Cross Correlation}
This metric has been used to establish the relation between geodesic and Euclidean distance in \cite{Brahma2016} for establishing the flattening achieved due to unsupervised pre-training of the deep network. While \cite{Hauberg_2017,Shao2018} discuss the degree of nonlinearity (or flatness) of the approximated data manifold in a VAE, the curvature of the established manifold is not quantified. The residual cross correlation measures the similarity between Euclidean and Riemannian distance and hence can provide an indirect measure of the curvature of the manifold. For example, for a linear manifold,  both the distances are equal, but nonlinearity or curvature of the space increases the difference between the two quantities. The residual cross correlation is given by 
\begin{align}
& c_k = 1- \frac{(r_{M}(k)-\mu_{r_M})(r_E(k)-\mu_{r_E})}{\sigma_{r_M}\sigma_{e_E}}\\
\nonumber &\hat{c} =\frac{2}{N(N-1)}\Sigma_{k}c_k
\end{align}
Here, $r_E$ and $r_M$ are the vectors formed by the concatenation of upper-triangular matrices of pairwise distance matrix for euclidean and manifold distance respectively. $\mu_{r_M}$, $\mu_{r_E}$ and $\sigma_{e_E}$, $\sigma_{e_M}$ are means and standard deviations for manifold and Euclidean distance respectively.
 \subsection{Normalized Margin}
 It measures the class separability in the latent space. A higher value of normalized margin means a larger separation between clusters of different class and hence a higher classification and clustering accuracy. Therefore, the specified space of disentangled representations would typically have a higher margin than a standard VAE.
 \begin{align}
 m_n = \frac{||x_n- \mathcal{M}(x_n)||-||x_n- \mathcal{H}(x_n)||}{||x_n- \mathcal{M}(x_n)||}
 \end{align}
Here, $\mathcal{M}(x_n)$ is the nearest member from a different class other than $x_n$ and $\mathcal{H}(x_n)$ is the nearest member from the same class. A larger value signifies  better separation between classes.

\subsection{Tangent Space Alignment}
For a flat manifold, the tangent spaces are aligned, so the angle between the two subspaces is zero. With the increase in the curvature of the space, the  tangent spaces at two points in this space would have a larger angle between them.

For a given data point $\bx$, the neighboring points are collected as $\bX=[\bx_1, \bx_2, \cdots \bx_k ]$. The mean centered data is given by 
\begin{align}
\hat{\bX}= \bX - \frac{1}{k}\bX \textbf{1}
\end{align}
The basis for the tangent space is obtained by the singular value decomposition of the covariance matrix given by 
\begin{align}
\bC = \bX^\top \bX =\bU \Sigma \bU^\top 
\end{align}
Given two points the principal angle between the subspaces defined by the tangent spaces gives a measure of curvature of the space.

\section{Experimental Setup and Results} \label{sec: exp}
In this section, we present the qualitative and quantitative results to analyze the geometric properties of the VAE and disentangled representation models.  We provide the details of the network architecture, datasets and the results for the different evaluation metrics.
\subsection{Datasets}
All the generative models are trained for 2 datasets: MNIST digits and MultiPIE face dataset. 
MNIST digits consist of 60000 training samples and 10000 test samples distributed over 10 class. The specified component is the class identity whereas the unspecified factors constitutes of digit slant, stroke width etc. We consider a subset of MultiPIE dataset with 25 identities for training and 5 identities for test, each with 3300 images. We use identity as specified factors and other factors like pose, expression etc are considered as unspecified. Further, we additionally evaluate the models on 3d chairs dataset that consists of images of different chair models of different styles and large viewpoint variations. We selected a subset of 30 models in  different viewing angles.
\subsection{Network Architecture}
We use Convolutional Neural Networks (CNNs) for all the models. The dimensions for the specified and unspecified spaces for MNIST digits are 16 and 64 respectively. For the Face dataset, the specified and unspecified dimensions are 512 and 64 respectively.For chair dataset, the specified and unspecified dimensions are  and respectively.  We also trained VAE with the same encoder decoder network as for disentangled representation models and with latent space dimensions as 16 and 512 for MNIST and Face dataset respectively. We used ELU activation function in all our models.
\subsection{Normalized Margin}
The normalized margin is a measure of distinguishibility of class specific features. The specified latent space of these models is enforced to learn class specific features and are used for task like classification. Classification accuracy is one measure to evaluate the effectiveness of these features for the given task. As pointed in \cite{Jha_2018}, the performance of the three models for MNIST digit classification is equally good, reflecting similar structure in the specified space of these models. The normalized margin value given in the Table \ref{tab: norm_margin} is consistent with the classification performance of these models. 
 \begin{table}[h!]
\centering
\begin{tabular}{|c|c|c|c|} 
 \hline
Datasets  & Szabo \emph{et al}& Mathieu \emph{et al.} & Jha \emph{et al.} \\\hline 
MNIST &  0.622 & 0.653 & 0.640 \\\hline
MultiPIE & 0.462 & 0.496 & 0.488 \\\hline
3D chairs & 0.492 & 0.529 & 0.52 \\\hline
 \end{tabular}
 \caption{Normalized Margin for MNIST, MultiPIE and 3D chairs dataset.}
 \label{tab:  norm_margin}
\end{table}
\subsection{Residual Cross Correlation $\hat{c}$} 
The values for the two datasets across different models are given in Table \ref{tab: c_value}. Smaller value suggests higher similarity between Euclidean and geodesic distances. For the MNIST dataset, the small value of $\hat{c}$ validates the claims of near zero curvature of VAEs discussed in \cite{Shao2018}. However, we observe that there is significant disparity between the Euclidean and the Riemannian distances for the MultiPie dataset. Note that Szabo et al. \cite{Szabo_2017} failed to converge on the MultiPie data, and hence are not included in the table. 
\begin{table}[h!]
\centering
\begin{tabular}{|c|c|c|c|c|}
\hline
Dataset & VAE & Szabo \emph{et al.}& Mathieu  \emph{et al.}& Jha \emph{et al.}\\\hline
MNIST & 0.071 & 0.142 & 0.178 & 0.167 \\\hline
MultiPie & 0.65 & - & 0.72 & 0.71 \\\hline
3D chairs & 0.162 & 0.262 & 0.311 & 0.315 \\\hline 
\end{tabular}
\caption{Comparison of $\hat{c}$ values for different disentangling models with VAE for MNIST digits, MultiPIE and 3D chairs dataset.}
\label{tab: c_value}
\end{table}
\vspace{-1cm}
\subsection{Curvature of Latent Spaces}
In this section, we evaluate the curvature of the latent space manifolds of VAEs. The low curvature claimed \cite{Shao2018} is reflected by the small residual cross correlation between Euclidean and Riemannian distance given in the Table \ref{tab: dim_curv} for different dimensions of the latent space. The stability of improvement in clustering performance across different dimensions is also indicative of the low curvature of the latent space regardless of the dimensionality.
We further quantify the curvature of latent spaces of VAE and disentangled  representation models in Table \ref{tab: angle} by computing the angle between the tangent spaces of a pair of points. The results validate higher curvature for disentangled representation models over VAEs. As the specified spaces of the three disentangled representation models are constrained to learn class discriminative features, the similarity in the curvature also reflects the same.
\begin{table}[h!]
\centering
\begin{tabular}{|c|c|c|c|} 
 \hline
 Dimension & 16 & 64 & 80 \\\hline 
 $\hat{c}$ & 0.065 & 0.069 & 0.071 \\\hline
F score (Euclidean)  & 83.32 & 85.22 & 87.38 \\\hline
 F score (Riemannian)  & 91.74 & 92.33 & 96.23 \\ \hline
 \end{tabular}
 \caption{Effect of dimensionality on the nonlinearity of the latent space of VAE . The clustering performance: F score with Euclidean and Riemannian distance as metric in K-means clustering algorithm for MNIST digits datasets.}
\label{tab: dim_curv}
\end{table}
 \begin{table}[htb!]
\centering
\small{
\begin{tabular}{|c|c|c|c|c|} 
 \hline
 & Models & VAE &  Mathieu \emph{et al.} & Jha \emph{et al.} \\\hline 
Distances & Euclidean & 0.312 & 0.346 & 0.332 \\\cline{2-5}
& Riemannian & 1.142 & 1.784 & 1.602 \\\hline
 Clustering & Euclidean & 82.98 & 89.37 & 90.06 \\\cline{2-5}
 F score & Riemannian & 89.04 & 94.45 & 95.60 \\\hline 
 \end{tabular}
 \caption{MultiPIE dataset: Comparison of Average distance between randomly selected 100 pairs and clustering performance performance: Riemannian Distance vs Euclidean Distance. The large differences in the distance/ F score is a result of curvature in the latent space.}
 \label{tab: r_e_face}
 }
\end{table}
\begin{figure}
	\begin{center}
		\includegraphics[width =0.49\textwidth]{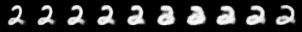}
        \includegraphics[width =0.49\textwidth]{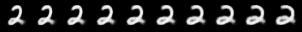}\\
        \end{center}
 \caption{Interpolation between two samples from same class in the latent space of VAE  using Euclidean (Top) and Riemannian Metric (Bottom).}
 \label{fig: centers_vae_same}
\end{figure}
\begin{figure}
	\begin{center}
         \includegraphics[width =0.45\textwidth]{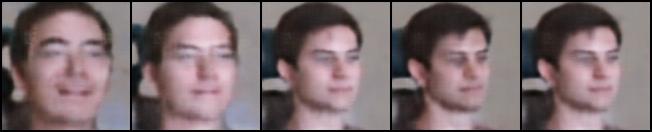}\\
        \includegraphics[width = 0.45\textwidth]{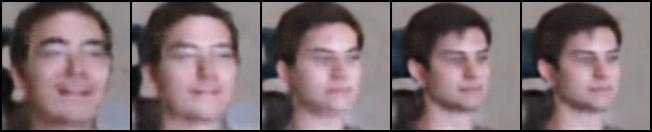}
        \\
        \;\;
		\includegraphics[width =0.45\textwidth]{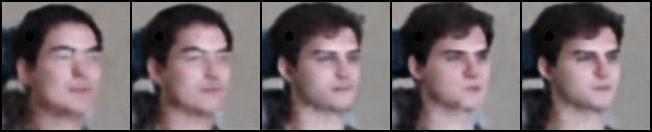}\\
         \includegraphics[width =0.45\textwidth]{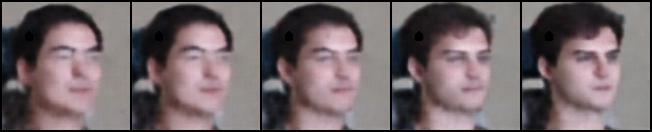}
        \end{center}
 \caption{Interpolation between two samples from different classes in the latent spaces of VAE (top) and specified space of  {Jha \emph{et al.}\cite{Jha_2018}} (bottom) with fixed unspecified using Euclidean ($1^{st}$ and $3^{rd}$ row) and Riemannian Metric ($2^{nd}$ and $4^{th}$ row).}
 \label{fig: face_inter}
\end{figure}
\begin{figure}
	\begin{center}
		\includegraphics[width =0.45\textwidth]{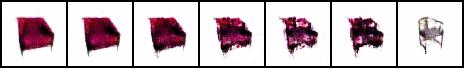} \includegraphics[width =0.45\textwidth]{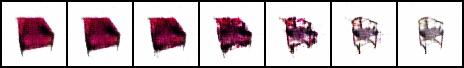} 
        \end{center}
 \caption{Interpolation between two samples from different classes in the latent spaces of {Mathieu \emph{et al.}\cite{Mathieu_2016}}  with fixed unspecified using Euclidean (top) and Riemannian Metric (bottom).}
 \label{fig: chair_example}
\end{figure}
\begin{figure*}
	\begin{center}
		\includegraphics[width =0.49\textwidth]{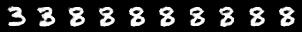}
        \includegraphics[width =0.49\textwidth]{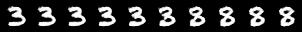}\\
        \includegraphics[width =0.49\textwidth]{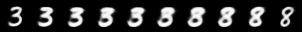}
        \includegraphics[width =0.49\textwidth]{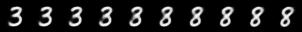}
        \end{center}
 \caption{Interpolation between two samples from different classes in the latent spaces of VAE and specified space of  {Mathieu \emph{et al.}\cite{Mathieu_2016}} with fixed unspecified using Euclidean (Left) and Riemannian Metric (Right).}
 \label{fig: centers_move_diff}
\end{figure*}
\begin{figure*}
	\begin{center}
		\includegraphics[width =\textwidth]{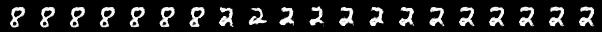} \includegraphics[width =\textwidth]{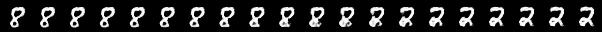}\\ 
        \;\;
        
        \includegraphics[width =\textwidth]{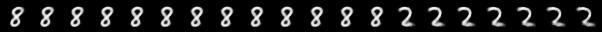}\\
        \includegraphics[width =\textwidth]{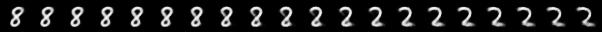}
        \end{center}
 \caption{Interpolation between two samples from different classes in the latent spaces of VAE (top) and specified space of  {Mathieu \emph{et al.}\cite{Mathieu_2016}} (bottom) with fixed unspecified using Euclidean ($1^{st}$ and $3^{rd}$ row) and Riemannian Metric ($2^{nd}$ and $4^{th}$ row).}
 \label{fig: centers_move_diff2}
\end{figure*}
\begin{figure}
	\begin{center}
		\includegraphics[width =0.49\textwidth]{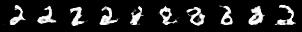}
        \includegraphics[width =0.49\textwidth]{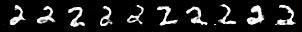}\\
        \end{center}
 \caption{Interpolation between  two samples from same class in the specified latent space of \textbf{Mathieu \emph{et al.}} with randomly sampled unspecified component using Euclidean (Top) and Riemannian Metric (Bottom).}
 \label{fig: centers_lecun_same}
\end{figure}
\begin{figure}
	\begin{center}
         \includegraphics[width =0.46\textwidth]{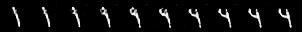}\\
        \includegraphics[width = 0.46\textwidth]{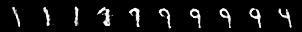}
        \\
        \;\;
		\includegraphics[width =0.46\textwidth]{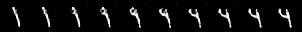}\\
         \includegraphics[width =0.46\textwidth]{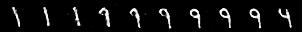}
        \end{center}
 \caption{Effect of ReLU (top) and ELU (bottom) activation functions on the quality of generated images with Euclidean ($1^{st}$ and $3^{rd}$ row) and Riemannian metric ($2^{nd}$ and $4^{th}$) interpolations.}
 \label{fig: relu_elu}
\end{figure}

 \begin{table}[h!]
\centering
\begin{tabular}{|c|c|c|c|c|} 
 \hline
Datasets  & VAE &  Szabo \emph{et al} & Mathieu \emph{et al.} & Jha \emph{et al.} \\\hline 
MNIST & 21.45 & 34.12 & 32.25 & 32.76 \\\hline
MultiPIE & 27.95 & 37.42 & 36.88 & 36.96 \\\hline
3D chairs &  23.45 & 36.77 & 35.86 & 36.50\\\hline
 \end{tabular}
 \caption{Approximate Curvature estimated with principal angles between Tangent Spaces.}
 \label{tab:  angle}
\end{table}
\subsection{Riemannian Distance vs Euclidean Distance}
While the $c$ values quantify the residual cross correlation between Euclidean and Riemannian distances, we also quote the magnitude of these distances in Table \ref{tab: r_e_mnist}, \ref{tab: r_e_chair} and  \ref{tab: r_e_face} for MNIST, 3D chairs and Face datasets respectively. Due to curvature of the specified space, using Riemannian distance over Euclidean distance is more appropriate for tasks like clustering as shown in the results. For both the datasets, a significant improvement in the clustering performance validates the  high curvature of the latent space for the disentangled models.
 \begin{table}[ht!]
\centering
\small{
\begin{tabular}{|c|c|c|c|c|} 
 \hline
 & Models & Szabo \emph{et al}& Mathieu \emph{et al.} & Jha \emph{et al.} \\\hline 
Distances  &Euclidean  & 0.114 & 0.112 & 0.116 \\\cline{2-5}
  &Riemannian & 0.297 & 0.355 & 0.336 \\\hline
 Clustering & Euclidean & 91.12 & 94.32 & 92.22 \\\cline{2-5}
  F score& Riemannian & 94.56 & 98.00 & 96.60 \\\hline 
 \end{tabular}
 \caption{MNIST dataset: Comparison of Average distance between randomly selected 100 pairs and clustering performance performance: Riemannian Distance vs Euclidean Distance. The large differences in the distance/ F score is a result of curvature in the latent space.}
 \label{tab: r_e_mnist}
 }
\end{table}
 \begin{table}[ht!]
\centering
\small{
\begin{tabular}{|c|c|c|c|c|} 
 \hline
 & Models & Szabo \emph{et al}& Mathieu \emph{et al.} & Jha \emph{et al.} \\\hline 
Distances  &Euclidean & 0.158 & 0.160 & 0.156 \\\cline{2-5}
  &Riemannian &  0.344 & 0.376 & 0.365 \\\hline
 Clustering & Euclidean & 91.16 & 94.33 & 94.24\\\cline{2-5}
  F score& Riemannian &  95.22 & 96.34 & 96.44 \\\hline 
 \end{tabular}
 \caption{3D chairs dataset: Comparison of Average distance between randomly selected 100 pairs and clustering performance performance: Riemannian Distance vs Euclidean Distance. The large differences in the distance/ F score is a result of curvature in the latent space.}
 \label{tab: r_e_chair}
 }
\end{table}
\vspace{-1cm}
\subsection{Interpolations}
Further, we also investigate the effect of Riemannian metric on interpolations in the latent space of VAEs as well as of the disentangling models. The images generated by linear and Riemannian interpolations for class 2 of MNIST for VAEs are shown in Figure \ref{fig: centers_vae_same}. The images generated with Riemannian metric are sharper as opposed to Euclidean metric reflecting the non-linear nature of the space. The two ends of the sequence are the images corresponding to the cluster center of class 2 and a point at the boundary of class 2 latent space representations. Thus along with sharper image generation, the presence of curvature in class specific manifold is also highlighted.  
For samples from different classes, the images for Euclidean and Riemannian interpolation are shown in Figure \ref{fig: centers_move_diff} for VAE as well as for the disentangling model. As shown in the figure, the transition from one class to other is abrupt with Euclidean distance while it changes more smoothly with the Riemannian counterpart. An example with large number of intermediate interpolants between class 8 and 2 is shown in the Figure \ref{fig: centers_move_diff2} for both VAE and disentangled representation model. The results  show the effect of the higher curved space in case of disentangled model with a huge difference between Riemannian and Euclidean interpolants as opposed to a VAE. Similar results are obtained for face and chair datasets and are shown in the Figure \ref{fig: face_inter} and \ref{fig: chair_example} respectively.
\subsection{Image Synthesis}
Disentangling specified component from other components, allows one to generate new images with different variations of fixed specified component. Figure \ref{fig: centers_lecun_same} shows the effect of using Euclidean and Riemannian  metric for generating images of a specific class with different styles. These images are obtained by interpolation in the specified space between the cluster center and a sample of the same class, while randomly sampling the unspecified component from Gaussian distribution imposed in the unspecified latent space. The samples generated are more realistic and preserve class identity in case of Riemannian metric.
\subsection{Rank of Jacobian}
For the network to learn smooth generator function, the Jacobian matrix is required to be full rank. We compare the rank of Jacobian matrix on MNIST dataset for the two activation functions ReLU and ELU. The results given in the Table \ref{tab: rank} show that the ELU results in full rank Jacobian matrices as opposed to ReLU. Further, the results in Figure \ref{fig: relu_elu} show the effect on the quality of the images generated with ReLU and ELU layers.  
\begin{table}[h!]
\centering
\begin{tabular}{|c|c|c|} 
 \hline
Activation function  & VAE &   Mathieu \emph{et al.} \\\hline 
ReLU & 8 & 15 \\\hline
ELU & 16 & 16\\\hline
 \end{tabular}
 \caption{Rank of the Jacobian matrix for MNIST digits.}
 \label{tab: rank}
\end{table}


\section{Conclusion}\label{sec: con}
Latent spaces of deep generative models provide a low dimensional representation for data embedded in a high dimensional manifold. In this work, we study the geometric properties of deep generative models that learn disentangled representations. We verified various recent claims about the nonlinearity of the VAE based latent space representations and utilized several metrics to quantify its lower curvature. Using the same metrics, we also established that the specified components of the latent space of VAE-based disentangled models is substantially more curved. The proposed study concludes that the latent spaces are curved, and thus an appropriate Riemannian metric as opposed to a Euclidean metric should be used for obtaining better distance estimates, performing interpolations as well as generating synthetic views. 

\section*{Acknowledgement}
This work is supported in part by Infosys Center for Artificial intelligence at IIIT Delhi, India.
\bibliographystyle{ACM-Reference-Format}
\bibliography{ICVGIP_main}

\end{document}